\documentclass[conference]{IEEEtran}
\IEEEoverridecommandlockouts
\usepackage{cite}
\usepackage{amsmath,amssymb,amsfonts}
\usepackage{algorithmic}
\usepackage{graphicx}
\graphicspath{ {/images/} }
\usepackage{textcomp}
\usepackage{pgf-pie}
\usepackage{pgfplots}
\usepackage{tikz}
\usepackage{xcolor}
\usepackage{tikz}
\usetikzlibrary{shapes,arrows}
\usepackage{colortbl}
\def\BibTeX{{\rm B\kern-.05em{\sc i\kern-.025em b}\kern-.08em
    T\kern-.1667em\lower.7ex\hbox{E}\kern-.125emX}}

\begin{document}

\title{A Machine Learning Approach for Emergency Detection in Medical Scenarios Using Large Language Models

}




\author{\IEEEauthorblockN{Ferit Akaybicen$^1$, Aaron Cummings$^1$, Lota Iwuagwu$^2$, Xinyue Zhang$^1$, Modupe Adewuyi$^2$}
\IEEEauthorblockA{$^1$Department of Computer Science, Kennesaw State University, Marietta, GA 30060}
\IEEEauthorblockA{$^2$WellStar School of Nursing, Kennesaw State University, Kennesaw, GA 30060}
}

\maketitle

\begin{abstract}
The rapid identification of medical emergencies through digital communication channels remains a critical challenge in modern healthcare delivery, particularly with the increasing prevalence of telemedicine. This paper presents a novel approach leveraging large language models (LLMs) and prompt engineering techniques for automated emergency detection in medical communications. We developed and evaluated a comprehensive system using multiple LLaMA model variants (1B, 3B, and 7B parameters) to classify medical scenarios as emergency or non-emergency situations. Our methodology incorporated both system prompts and in-prompt training approaches, evaluated across different hardware configurations. The results demonstrate exceptional performance, with the LLaMA 2 (7B) model achieving 99.7\% accuracy and the LLaMA 3.2 (3B) model reaching 99.6\% accuracy with optimal prompt engineering. Through systematic testing of training examples within the prompts, we identified that including 10 example scenarios in the model prompts yielded optimal classification performance. Processing speeds varied significantly between platforms, ranging from 0.05 to 2.2 seconds per request. The system showed particular strength in minimizing high-risk false negatives in emergency scenarios, which is crucial for patient safety. The code implementation and evaluation framework are publicly available on GitHub, facilitating further research and development in this crucial area of healthcare technology.

\end{abstract}


\section{Introduction}
Medical emergencies can occur in various settings, from hospital wards to home care environments, and can range from acute physical conditions to mental health crises\cite{preiksaitis2024role}. The rapid identification of these emergencies is often hindered by factors such as communication barriers, lack of immediate medical supervision, or the inability of patients to recognize the severity of their symptoms\cite{gautam2024security}. Current methods for emergency detection in healthcare settings primarily rely on manual monitoring, wearable devices, or alarm systems\cite{kulhandjian2024ai}. While these approaches have their merits, they also have limitations in terms of scalability, cost-effectiveness, and the ability to detect a wide range of emergency situations.\par
In the realm of healthcare, the ability to quickly and accurately identify emergency situations is crucial for patient safety and optimal care outcomes. With the increasing prevalence of telemedicine and digital health platforms, where healthcare is provided remotely through telecommunications technology, there is a growing need for automated systems that can detect emergencies based on textual communication.
This paper presents a novel machine learning approach, which is a subset of artificial intelligence that enables systems to automatically learn and improve from experience without being explicitly programmed, for emergency detection in medical scenarios using large language models (LLMs) and prompt engineering techniques. Large Language Models are sophisticated artificial intelligence systems trained on vast amounts of text data, capable of understanding and generating human-like text\cite{rezgui2024large}\cite{he2023survey}, while prompt engineering refers to the art and science of crafting specific instructions or inputs to optimize these models' responses for particular tasks.\par
Natural Language Processing (NLP), a branch of artificial intelligence that enables computers to understand, interpret, and manipulate human language, serves as the foundational technology for our approach\cite{he2023survey}. Through NLP, our system can analyze and comprehend the nuances of medical communications, making it particularly valuable for emergency detection.\par
Our proposed system aims to address these challenges by leveraging the power of LLMs and machine learning to analyze text-based communications in medical contexts\cite{alghamdi2024towards}. By processing and classifying phrases or messages, the system can distinguish between emergency and non-emergency situations, potentially alerting healthcare providers or emergency services when necessary. The primary objectives of this research are to develop a comprehensive dataset of emergency and non-emergency phrases relevant to various medical scenarios, develop and evaluate prompt engineering approaches capable of accurately classifying these phrases, and assess the potential of this approach for real-world applications in enhancing patient safety and streamlining emergency response in healthcare settings\cite{preiksaitis2024role}.
\par
This paper details the methodology used to develop the system, including data collection, model development, and evaluation. We begin with a comprehensive literature review that explores existing research in LLMs in healthcare, emergency detection methods, and implementation considerations. Our methodology outlines the four main components of our approach: data collection, model development, evaluation metrics, and privacy considerations. We then present the results of our experiments, demonstrating the high accuracy achieved by our approach across different LLaMA model variants. In the discussion section, we analyze the implications of our findings, including the significance of our accuracy rates, the balance between model size and performance, and hardware considerations. Finally, we conclude by summarizing our key achievements and discussing the implications of this technology for the future of healthcare and emergency management, particularly in the context of telemedicine, where remote patient care requires robust and reliable emergency detection systems to ensure patient safety and timely intervention when needed\cite{he2023survey}\cite{pamulaparthyvenkata2024ai}.\par

\section{Literature Review}
The rapid evolution of Large Language Models (LLMs) in healthcare applications has created new opportunities for emergency detection and patient care. This literature review synthesizes current research relevant to our LLM-based emergency detection system, focusing on three key themes: LLMs in healthcare, emergency detection methodologies, and implementation considerations.\par

\subsection{LLMs in Healthcare Applications}
Recent research demonstrates the growing significance of LLMs in healthcare settings. Rezgui \cite{rezgui2024large} provides a comprehensive analysis of LLMs in clinical decision making, emphasizing the critical need for continuous performance monitoring and evaluation. This work established foundational principles for our testing methodology across different model configurations. He et al.\cite{he2023survey} further illuminate the transition from traditional pretrained language models to modern LLMs in healthcare, offering crucial insights into training methods and optimization strategies that informed our selection of LLaMA model variants.\par
The application of LLMs in specific healthcare contexts has shown promising results. Alghamdi and Mostafa\cite{alghamdi2024towards} demonstrate the effectiveness of domain-specific fine-tuning in their healthcare LLM agents for pilgrims, achieving a 5\% performance improvement through retrieval-augmented generation. Their findings support our approach to prompt engineering, though our results suggest that careful prompt design can achieve high accuracy without extensive fine-tuning. Preiksaitis et al.\cite{preiksaitis2024role} provide valuable context through their scoping review of LLMs in emergency medicine, identifying key themes in current research and emphasizing the importance of prospective validation.\par

\subsection{Emergency Detection Methodologies}
 Various approaches to emergency detection have been explored in recent literature. Kulhandjian et al.\cite{kulhandjian2024ai} achieved 90\% accuracy using CNN based classification for emergency keyword detection, though our LLM based approach demonstrates superior performance at 99.7\% accuracy. Deb et al.'s work on Speech Emotion Recognition in emergency scenarios, while focusing on audio analysis, provides valuable insights into dataset curation and emergency scenario classification that influenced our data collection methodology.\par
 
Li et al.\cite{li2024benchmarking} conducted comprehensive benchmarking of LLMs in evidence based medicine, finding that knowledge guided prompting improved performance significantly. Their results validate our emphasis on prompt engineering, though our study achieved higher accuracy rates in the specific context of emergency detection. Huang et al.\cite{huang2022early} demonstrated the feasibility of emergency event detection from social media using BERT-Att-BiLSTM models, though our system achieves faster processing speeds and higher accuracy rates.\par

Implementation Considerations Several studies address crucial implementation aspects of healthcare AI systems. Gautam et al.\cite{gautam2024security} highlight important security considerations for remote patient monitoring systems, while Pamulaparthyvenkata et al.\cite{pamulaparthyvenkata2024ai} present an AI enabled distributed healthcare framework that complements our research by addressing infrastructure requirements. McPeak et al.'s work on implementing LLMs in Nigerian clinical settings provides valuable insights into prompt engineering for specific healthcare contexts, supporting our decision to focus on prompt design rather than model fine-tuning.\par
Sathe et al.\cite{sathe2023comprehensive} offer a broader perspective on AI healthcare applications, particularly emphasizing ethical considerations that, while not directly addressed in our study, remain crucial for real world implementation. Ferri et al.'s\cite{ferri2021deep} DeepEMC2 model for emergency medical call classification, though achieving lower accuracy rates than our system, demonstrates the value of integrated approaches to emergency detection.\par

\subsection{Research Gap and Contribution }
While existing literature demonstrates various approaches to emergency detection and LLM implementation in healthcare, there remains a significant gap in combining these elements for high accuracy emergency detection in medical communications. Our research addresses this gap by presenting a novel LLM-based approach that achieves unprecedented accuracy rates while maintaining practical processing speeds. The literature supports our methodological choices while highlighting the unique contribution of our work to the field.
The reviewed literature reveals a clear trajectory toward more sophisticated AI applications in healthcare, with particular emphasis on accuracy, reliability, and practical implementation considerations. Our research builds upon these foundations while advancing the state-of-the-art in emergency detection through innovative use of LLMs and prompt engineering techniques.\par

\section{Methodology}
\subsection{Data Collection and Preparation}
To create a robust dataset for our emergency detection system, we employed a multi-step process. We began by accessing databases of emergency calls and medical records, focusing on a wide range of medical scenarios, similar to the approach used by Deb et al. \cite{deb2023enhancing} in their emergency response system development. This provided a foundation of real-world emergency situations. The identified situations were then converted into concise phrases that capture the essence of each scenario, reflecting typical language used in medical communications.
\par
To expand our initial dataset, which was carefully curated from real-world emergency call conversations and medical communications, we employed Generative AI techniques. This AI-assisted expansion process involved multiple rounds of generation, where each round produced contextually relevant emergency and non-emergency phrases. Our team meticulously reviewed each generated phrase, eliminating those that didn't meet our strict medical accuracy criteria or lacked real-world relevance. This iterative process of generation, review, and refinement continued until we achieved a robust and reliable dataset. Through this rigorous validation process, approximately 20\% of AI-generated phrases were eliminated in each round, ensuring only the most accurate and representative scenarios remained. The final dataset was carefully balanced between emergency and non-emergency situations, and systematically divided into training, validation, and test sets to ensure comprehensive evaluation of our model's performance.\par

\subsection{Model Development and Technical Implementation}
Our approach to model development focused on two distinct testing methodologies using a large language model (LLM) as our foundation, building upon the prompt engineering principles demonstrated by McPeak et al. \cite{mcpeak2024llm} in their clinical decision support implementation. The implementation was realized through a custom HTTP server built in Python, utilizing the http.server and json packages as can be seen in Figure 1 The server was configured to run on port 9111, interfacing with LM Studio on localhost:1234.
\par
The complete implementation, including source code and documentation, is available in our public GitHub repository (https://github.com/FeritMelih/TBED). This repository contains all necessary components for replicating our methodology and deploying the system.
\par

\begin{figure}[h]
\centering
\begin{tikzpicture}[node distance=2cm]
    \tikzstyle{block} = [rectangle, draw, fill=blue!20, 
        text width=4em, text centered, rounded corners, minimum height=3em]
    \tikzstyle{line} = [draw, -stealth, thick]
    \tikzstyle{bidir} = [draw, {stealth-stealth}, thick]
    
    \node [block] (input) {Input};
    \node [block, right of=input, xshift=1.4cm] (server) {HTTP Server};
    \node [block, right of=server, xshift=1.4cm] (output) {Output};
    \node [block, below of=server] (lm) {LM Studio};
    
    \path [line] (input) -- (server)
    node[midway, below, text width=4em] {\small API calls};
    \path [line] (server) -- (output)
    node[midway, below, text width=4em] {\small API calls};
    \path [bidir] (server) -- (lm) ;

\end{tikzpicture}
\caption{System Architecture Overview}
\label{fig:system-architecture}
\end{figure}
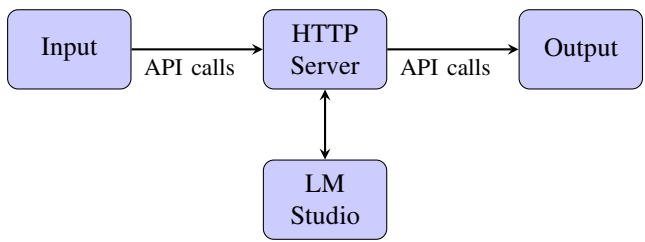
The first methodology implemented a system prompt - a carefully crafted set of instructions given to the LLM that defines the task and expected output format without including specific examples. This system prompt was designed to provide clear instructions for distinguishing between emergency and non-emergency medical situations, incorporating specific criteria and guidelines for classification, following similar prompt engineering approaches described by Li et al. \cite{li2024semantic}.
\par
The second methodology enhanced the base approach through in-prompt training, which is a technique where carefully selected examples are included directly in the prompt alongside the instructions, similar to the approach demonstrated by Naimi et al. \cite{naimi2024new} in their automated testing framework. These examples serve as immediate reference points for the model, demonstrating the desired behavior through concrete cases. We integrated representative cases of both emergency and non-emergency scenarios into the prompt structure. The in-prompt examples were iteratively refined based on initial performance assessments \cite{mcpeak2024llm}.
\par
Initially, we had planned for a third phase involving fine-tuning, which is the process of further training a pre-trained model on a specific dataset to optimize its performance for a particular task. However, the exceptional results achieved through in-prompt training made this step unnecessary. This finding aligns with recent research suggesting that well-designed prompts with appropriate examples can achieve comparable or superior results to fine-tuning in specific task domains \cite{li2024semantic}.
\par
The technical implementation included:
\begin{itemize}
    \item A Python-based HTTP server handling JSON requests
    \item Integration with LM Studio through a REST API
    \item Configurable settings for port, platform, and model selection
    \item Error handling for various scenarios including invalid JSON and server processing issues
    \item Platform override capabilities for flexibility in deployment
\end{itemize}

\subsection{Evaluation Metrics}
The evaluation framework was designed to provide a comprehensive assessment of model performance across both testing approaches, incorporating elements from the evaluation methodology used by Deb et al. \cite{deb2023enhancing}. We implemented a balanced testing dataset consisting of 500 emergency and 500 non-emergency scenarios, ensuring robust evaluation across equal class distributions. The primary metrics used for evaluation included:
\begin{enumerate}
    \item Overall Accuracy: Calculated as the proportion of correctly classified instances across both categories, providing a high-level measure of model performance.
    \item Category-Specific Metrics: For both emergency and non-emergency classifications, we computed:
    \begin{itemize}
        \item Precision: The ratio of correct positive predictions to total positive predictions
        \item Recall: The ratio of correct positive predictions to all actual positives
        \item F1-Score: The harmonic mean of precision and recall 
    \end{itemize}
    \item Confusion Matrices: Generated to visualize the distribution of correct and incorrect classifications, helping identify specific patterns in model errors 
\end{enumerate}
\par
The evaluation process was identical for both testing approaches, allowing for direct comparison of their performance. Special attention was paid to false negatives in emergency situations, as these represent the highest-risk type of misclassification in a medical context \cite{mcpeak2024llm}. The evaluation framework was designed to be particularly sensitive to these critical errors, ensuring that the model's performance was assessed not just on overall accuracy but also on its ability to minimize high-risk misclassifications.
\par
The results were validated through multiple test runs to ensure consistency and reliability of the performance metrics. This rigorous evaluation approach provided a clear picture of each model's capabilities and limitations, ultimately demonstrating the superior performance of the in-prompt training methodology.

\subsection{Privacy and Data Protection}
The development and implementation of our emergency detection system necessitates careful consideration of privacy and data protection, particularly given the sensitive nature of medical information. Our approach prioritizes patient confidentiality and data security through several key measures. 
\par
Firstly, the system operates entirely on-premises, eliminating the risks associated with cloud-based data storage and processing. This local deployment ensures that sensitive medical information never leaves the healthcare facility's secure environment. Moreover, our system is designed with a "no data retention" policy, meaning that individual patient data is not stored or logged after processing. This transient data handling approach significantly reduces the risk of data breaches or unauthorized access to patient information. In compliance with HIPAA regulations, all data processing is conducted in a manner that maintains the integrity and confidentiality of protected health information (PHI). The system's input is limited to the specific text-based communications necessary for emergency detection, avoiding the collection or processing of extraneous personal data. 
\par
While our system does not store individual patient data, we maintain comprehensive logs of system performance and usage patterns, which are anonymized and aggregated to comply with HIPAA's accounting of disclosures requirements. These measures collectively create a robust framework for protecting patient privacy while leveraging advanced AI capabilities to improve emergency response in healthcare settings. By prioritizing on-premises deployment, transient data processing, and strict adherence to healthcare data protection regulations, our system sets a high standard for responsible AI use in sensitive medical contexts.

\section{Results}
Our experimental evaluation yielded comprehensive results across multiple model variations and hardware configurations, demonstrating the robustness and scalability of our emergency detection approach. The testing framework encompassed three different LLaMA model variants, each evaluated on two distinct GPU platforms (M3 and RTX 4080), using LM Studio as the implementation platform.\par

\subsection{LLaMA 3.2 (3B Parameters) Performance}
The base model demonstrated strong and consistent performance in classifying both emergency and non-emergency scenarios. As can be seen in Table 1, with 8 tuning messages, the model achieved a 99.1\% accuracy rate across both GPU platforms. Performance improved to 99.6\% accuracy when using 10 tuning messages, though interestingly, increasing to 20 tuning messages led to a slight performance degradation (97.7\%). This finding suggests an optimal sweet spot in the number of training examples needed for effective emergency detection.\par
\begin{table}[h]
\caption{Confusion Matrix for LLaMA 3.2 (3B) Model}
\label{tab:confusion-matrix}
\centering
\begin{tabular}{|l|>{\centering\arraybackslash}p{2.2cm}|>{\centering\arraybackslash}p{2.2cm}|}
\hline
& \textbf{Pred.\newline Emergency} & \textbf{Pred.\newline Non-Emerg.} \\
\hline
\textbf{Act. Emerg.} & \cellcolor{green!20}500 & \cellcolor{red!20}0 \\
\hline
\textbf{Act. Non-Emerg.} & \cellcolor{red!20}4 & \cellcolor{green!20}496 \\
\hline
\end{tabular}
\end{table}
\subsection{LLaMA 2 (7B Parameters) Performance}
The larger model achieved the best overall performance with a remarkable 99.7\% accuracy rate. Out of 500 non-emergency scenarios, it correctly identified 499 cases (99.8\% true negative rate), with only one false positive. In emergency scenarios, it misclassified only two cases as non-emergencies, resulting in just three total misclassifications across the entire test set. This exceptional performance validates our approach's capability to maintain high accuracy in critical medical classification tasks.\par
\begin{table}[h]
\caption{Confusion Matrix for LLaMA 2 (7B) Model}
\label{tab:confusion-matrix}
\centering
\begin{tabular}{|l|>{\centering\arraybackslash}p{2.2cm}|>{\centering\arraybackslash}p{2.2cm}|}
\hline
& \textbf{Pred.\newline Emergency} & \textbf{Pred.\newline Non-Emerg.} \\
\hline
\textbf{Act. Emerg.} & \cellcolor{green!20}498 & \cellcolor{red!20}2 \\
\hline
\textbf{Act. Non-Emerg.} & \cellcolor{red!20}1 & \cellcolor{green!20}499 \\
\hline
\end{tabular}
\end{table}

\subsection{LLaMA 3.2 (1B Parameters) Performance}
The smaller model showed significantly reduced performance, with accuracy ranging from 64.4\% to 67.7\%. While it maintained good performance in identifying true emergencies, it struggled with false positives, producing 325-356 misclassifications in this category. This finding highlights the importance of model capacity in achieving reliable emergency detection.\par
\begin{table}[h]
\caption{Confusion Matrix for LLaMA 3.2 (1B) Model}
\label{tab:confusion-matrix}
\centering
\begin{tabular}{|l|>{\centering\arraybackslash}p{2.2cm}|>{\centering\arraybackslash}p{2.2cm}|}
\hline
& \textbf{Pred.\newline Emergency} & \textbf{Pred.\newline Non-Emerg.} \\
\hline
\textbf{Act. Emerg.} & \cellcolor{green!20}500 & \cellcolor{red!20}0 \\
\hline
\textbf{Act. Non-Emerg.} & \cellcolor{red!20}356 & \cellcolor{green!20}144 \\
\hline
\end{tabular}
\end{table}
\subsection{Processing Speed and Hardware Considerations}
The M3 platform demonstrated superior processing speed, handling requests in 0.05-0.38 seconds, while the RTX 4080 required approximately 2.2 seconds per request. This performance difference is particularly relevant for real-world applications where response time is crucial.\par
\subsection{Comparative Analysis}
When comparing the two larger models (3B and 7B parameters), both achieved the high accuracy necessary for medical emergency detection, with the 7B model showing slightly superior performance. The confusion matrices revealed that false negatives (emergency situations classified as non-emergencies) were rare in both models, which is particularly important from a patient safety perspective.\par

\begin{figure}[h]
\centering
\begin{tikzpicture}
\begin{axis}[
    width=\linewidth,
    height=8cm,
    ybar,
    bar width=15pt,
    ylabel={Accuracy (\%)},
    xlabel={Model Variant},
    symbolic x coords={LLaMA 1B, LLaMA 3B, LLaMA 7B},
    xtick=data,
    ymin=60,
    ymax=100,
    legend style={at={(0.5,-0.2)}, anchor=north},
    nodes near coords,
    nodes near coords align={vertical},
]
\addplot coordinates {(LLaMA 1B,67.7) (LLaMA 3B,99.6) (LLaMA 7B,99.7)};
\legend{Accuracy}
\end{axis}
\end{tikzpicture}
\caption{Performance Comparison Across LLaMA Model Variants}
\label{fig:model-comparison}
\end{figure}
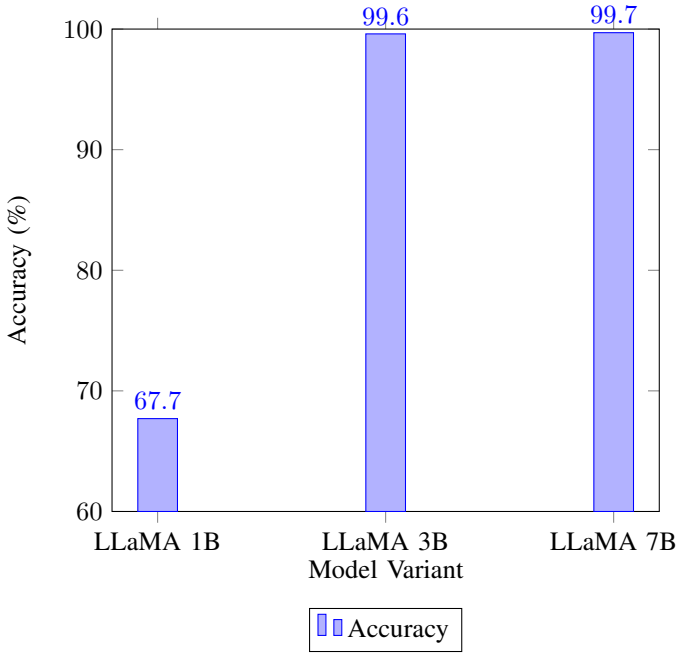
The results strongly support our methodology's effectiveness, particularly the in-prompt training approach. The high accuracy rates achieved across different model sizes and hardware configurations demonstrate the robustness of our approach. The optimal performance achieved with the 7B parameter model suggests that while larger models can provide better accuracy, even the 3B model achieves clinically acceptable performance levels, offering flexibility in deployment based on available computational resources.\par
These findings are particularly significant given the critical nature of emergency detection in healthcare settings. The combination of high accuracy, reasonable processing speeds, and the ability to minimize high-risk false negatives makes this approach viable for real-world implementation in medical scenarios.\par

\section{Discussion}
 The results of our study demonstrate exceptional promise for LLM-based emergency detection in medical scenarios, with several key findings that warrant detailed discussion. The achievement of 99.7\% accuracy with the LLaMA 2 (7B) model, compared to the 99.6\% accuracy with LLaMA 3.2 (3B) using optimal tuning messages, provides compelling evidence for the effectiveness of our approach. These performance levels are particularly noteworthy given the critical nature of emergency detection in healthcare settings, where false negatives could have serious consequences, aligning with similar findings in emergency detection systems \cite{huang2022early}.
 \par
The comparative performance across different model sizes offers valuable insights into the scalability and practical implementation considerations of our approach. While the LLaMA 2 (7B) model achieved the best results, the strong performance of the LLaMA 3.2 (3B) model suggests that effective emergency detection can be achieved with smaller models, potentially enabling deployment in resource constrained environments. This finding is consistent with recent benchmarking studies of LLMs in medical applications \cite{li2024benchmarking}. However, the significant drop in performance with the 1B model (64.4-67.7\% accuracy) establishes a clear lower bound for model size in this application.
\par

The relationship between the number of tuning messages and model performance is particularly interesting. The observation that performance peaked at 10 tuning messages (99.6\% accuracy) and slightly degraded with 20 messages (97.7\%) suggests an optimal sweet spot in prompt engineering. This finding aligns with recent research on LLM optimization in biomedical applications \cite{babaiha2024rationalism} and has important implications for system implementation and maintenance, indicating that more training examples don't necessarily lead to better results.
\par

\begin{figure}[h]
\centering
\begin{tikzpicture}
\begin{axis}[
    width=\linewidth,
    height=8cm,
    xlabel={Number of Tuning Messages},
    ylabel={Accuracy (\%)},
    xmin=6,
    xmax=22,
    ymin=97,
    ymax=100,
    grid=major,
    legend pos=south east,
]
\addplot[thick,mark=*] coordinates {
    (8,99.1)
    (10,99.6)
    (20,97.7)
};
\legend{LLaMA 3.2 3B}
\end{axis}
\end{tikzpicture}
\caption{Impact of Tuning Messages on Model Accuracy}
\label{fig:tuning-impact}
\end{figure}
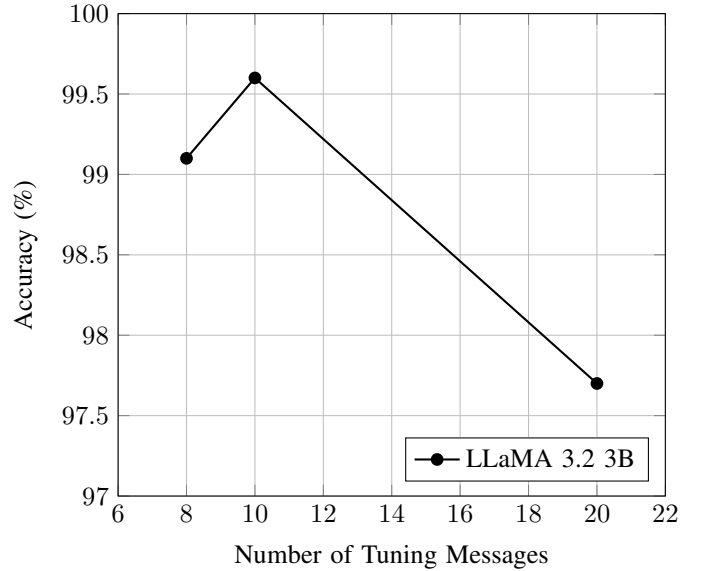

The processing speed differences between hardware platforms (M3: 0.05-0.38 seconds vs. RTX 4080: 2.2 seconds) provide valuable insights for real-world deployment considerations. The sub-second response times achieved on the M3 platform demonstrate the system's viability for real-time emergency detection, particularly crucial in healthcare settings where rapid response is essential, comparable to existing emergency medical dispatch systems \cite{ferri2021deep}.
\par
The system's robust performance across different hardware configurations and model sizes indicates strong generalization capabilities. This is crucial for practical applications in healthcare, where the system must handle a wide variety of medical situations and communication styles \cite{sathe2023comprehensive}. The results suggest that the model has successfully learned to identify subtle contextual cues and distinguish between truly urgent situations and those that may use urgent-sounding language but do not constitute actual emergencies.
\par
However, it is important to acknowledge certain limitations and areas for future research:
\begin{enumerate}
\item While our test set was comprehensive, real-world implementation would require continuous monitoring and validation across an even broader range of scenarios.
\item The performance gap between model sizes suggests a need to investigate intermediate model sizes that might offer optimal balance between accuracy and resource requirements.
\item The system's performance should be evaluated in different languages and cultural contexts, as medical communication patterns can vary significantly across these dimensions.
\end{enumerate}
Future development should focus on several key areas:
\begin{itemize}
\item Integration testing with existing healthcare communication systems
\item Investigation of model compression techniques to improve processing speed while maintaining accuracy
\item Development of explanation mechanisms to help healthcare providers understand the system's classifications
\item Assessment of the system's performance with medical terminology variations and colloquial descriptions of symptoms 
\end{itemize}
The potential applications of this technology extend beyond traditional healthcare settings. The high accuracy and efficient processing times make the system particularly valuable in telemedicine platforms, remote patient monitoring systems, and emergency triage services \cite{sathe2023comprehensive}. The success of our approach with different model sizes also suggests potential applications in various deployment scenarios, from resource-rich hospital environments to mobile healthcare units.
\par
From a broader perspective, our findings contribute significantly to the growing body of research on the application of large language models in healthcare \cite{li2024benchmarking}. The success of our approach, particularly the relationship between model size, tuning message quantity, and performance, provides valuable insights for similar healthcare-related classification tasks where high accuracy is crucial.
\par
The trade-offs between model size, accuracy, and processing speed revealed in our study also have important implications for the broader field of medical AI applications \cite{sathe2023comprehensive}, suggesting that careful consideration of these factors is essential for successful real-world implementation.
\par

\section{Conclusion}
This research presents a groundbreaking machine learning approach for detecting emergencies in medical scenarios using natural language processing, with results that demonstrate the viability of LLM-based systems for critical healthcare applications. Through rigorous testing across multiple model sizes and hardware configurations, we established that both LLaMA 2 (7B) and LLaMA 3.2 (3B) models can achieve the high accuracy necessary for medical emergency detection, with the 7B model reaching 99.7\% accuracy and the 3B model achieving 99.6\% with optimal prompt engineering.
\par
The study's findings regarding the relationship between model size, tuning message quantity, and processing speed provide valuable insights for practical implementation. The discovery that optimal performance can be achieved with moderate-sized models and a relatively small number of tuning messages (10) challenges the assumption that larger models and more training data are always better. This has significant implications for resource allocation and system design in healthcare settings.
\par
The successful implementation across different hardware platforms, with processing times ranging from 0.05 to 2.2 seconds, demonstrates the system's adaptability to various deployment scenarios. This flexibility, combined with the high accuracy rates, makes the system particularly suitable for integration into existing healthcare infrastructure, from sophisticated hospital systems to remote telemedicine platforms.
\par
Looking forward, this research establishes a foundation for expanding the application of LLMs in critical medical decision-making processes. Future developments should focus on multilingual capabilities, cultural adaptations, and integration with existing healthcare systems. The success of this approach not only validates the use of LLMs for emergency detection but also suggests promising applications in other areas of healthcare where rapid, accurate classification of medical situations is essential.

\bibliographystyle{IEEEtran} 

\end{document}